\documentclass[8pt,twocolumn]{article}
\usepackage[utf8]{inputenc}
\usepackage{amsmath,amssymb}
\usepackage{fullpage,hyperref}
\usepackage{verbatim}
\usepackage{multimedia}
\usepackage{graphicx}
\def\bt{{\bf t}}
\title{Ancestral Inference from Functional Data: Statistical Methods and Numerical Examples} 
\author{Pantelis Z. Hadjipantelis \footnote{Centre for Complexity Science, University of Warwick, Coventry CV4 7AL, UK}, Nick, S. Jones\footnote{Department of Mathematics, Imperial College London, SW7 2AZ}, John Moriarty\footnote{School of Mathematics, University of Manchester, Oxford Road, Manchester M13 9PL, UK} , David Springate\footnote{Faculty of Life Sciences, University of Manchester, Oxford Road, Manchester M13 9PT, UK} , Christopher G. Knight\footnotemark[\value{footnote}]}
\date{}
 
\begin{document}
\maketitle

\begin{abstract}
Many biological characteristics of evolutionary interest are not scalar variables but continuous functions. Here we use phylogenetic Gaussian process regression to model the evolution of simulated function-valued traits. Given function-valued data only from the tips of an evolutionary tree and utilising independent principal component analysis (IPCA) as a method for dimension reduction, we construct distributional estimates of ancestral function-valued traits, and estimate parameters describing their evolutionary dynamics. \end{abstract} 


\section{Introduction}
\label{sec:Intro}

The number, reliability and coverage of evolutionary trees is growing rapidly \cite{Maddison07}.However, knowing organisms' evolutionary relationships through phylogenetics is only one step in understanding the evolution of their characteristics \cite{Yang12}. Two issues are particularly challenging: first, information is typically only available for extant organisms, represented by tips of a phylogenetic tree, whereas understanding their evolution requires inference about ancestors deeper in the tree. Second, available information for different organisms in a phylogeny is not  independent: a phylogeny describes a complex pattern of non-independence.  
A variety of statistical models have been developed to address these issues (e.g. see \cite{Salamin10}). However, most deal with only one characteristic of an organism, encapsulated in a single value, at a time. This simplicity contrasts with the complexity of any organism which has, not only many individual characteristics, but also characteristics impossible to  represent effectively as single numbers. Some such characteristics, for example growth curves, may be modelled as \emph{function-valued traits} \cite{Kirkp89}, i.e. as points in an infinite dimensional space. 
A novel approach to analysing the evolution of function-valued traits has recently been proposed: phylogenetic Gaussian process regression \cite{JM12, Funcphylo11}.
Here we develop a practical implementation of this approach, which first linearly decomposes function-valued traits for a set of taxa into statistically independent components. This phylogenetically agnostic method of dimension reduction is robust to mixed inherited and taxon-specific variation in the data (e.g. see \cite{Cheverud85}).  
 Our method then implements the Bayesian regression analysis described in \cite{JM12}, returning posterior distributions for ancestral function-valued traits. We show that this analysis produces reliable distributional estimates for our simulated data, which may be further separated into inherited and specific components of variation. We also analyse the statistical performance of the method for making point estimates for the (hyper-)parameters \cite{Rasmussen06} describing the evolutionary dynamics of the components. Overall, our method appropriately combines developments in functional data analysis with the evolutionary dynamics of quantitative phenotypic traits, allowing nonparametric Bayesian inference from phylogenetically non-independent function-valued traits.  
Details of the simulation and inference methods are given in section \ref{sec:Methods} and section  \ref{sec:ResDes} gives results and discussion both for reconstructing ancestral function-valued traits and for estimating their evolutionary parameters.
 
\section{Methods}
\label{sec:Methods}

\subsection{Simulating function-valued traits}
\label{sec:Simulation} 
We simulate datasets using the Ornstein-Uhlenbeck (OU) Gaussian process as a model of evolutionary change. The technical justification for the broad applicability of this model is presented in \ref{sec:DimensionRed} and \ref{sec:PhyloGP}.  
However, OU processes are already well documented in the evolutionary biology literature \cite{Butler04,Hansen96}, having the advantage over simpler Brownian motion models \cite{Felse85} of modelling both selection and genetic drift. The OU model also exhibits a stationary distribution with covariances between character values decreasing exponentially with phylogenetic distance \cite{Hanse97}.

\begin{flushleft}\begin{figure*}[!ht]
\includegraphics[width=1.0\textwidth]{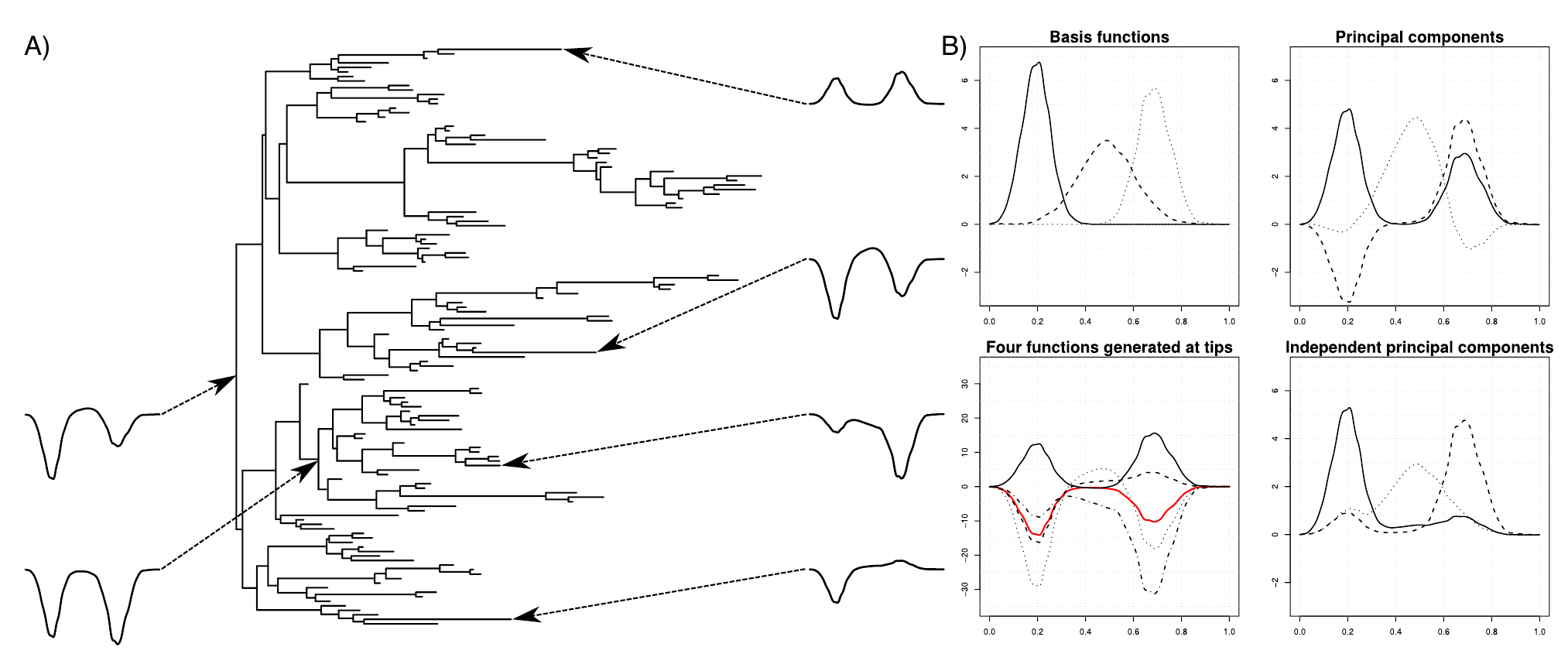}
\caption{$\bf{A}$ The 128-tip random tree used with function-valued traits at four tips (right), the root and one internal node (left). The maximal distance between any two tips is 17.9, the mean distance 7.43. $\bf{B}$ Panels, clockwise from top-left: The common basis functions used to generate data ($\phi_1, \phi_2, \phi_3$); results of PCA on the training data; results of IPCA on the training data; and the simulated function-valued traits at the four tips indicated in $\bf{A}$ (black lines), with the sample mean curve (red). } \label{FIG1}
\end{figure*}\end{flushleft}

We first generated a random, non-ultrametric, 128-tip phylogentic tree $\bf{T}$, with branch lengths drawn from an inverse Gaussian (IG) distribution, IG(.5,.5) (figure \ref{FIG1}A).  
Function-valued traits were simulated at each tip and internal node by randomly mixing a common set of basis functions; for each \textit{i }= 1, 2, 3 a smooth function (figure 1B,Upper-left) was discretised producing a basis vector $\phi_i$ of length 1024. An OU Gaussian Process (independent for each $i$) was used to generate weights for the $i$th component $w^i$ at each of the 255 nodes of ${\bf T}$ (128 tips and 127 internal nodes), with mean zero and covariance function
\begin{eqnarray}
&& k_{\bf T}^i(\bt_1,\bt_2) = E[w^i_{\bt_1},w^i_{\bt_2}] \label{oucov}\\ && \quad = (\sigma_f^{i})^2 exp\left( \frac{-d_T(\bt_1,\bt_2)}{\lambda^i}\right) + (\sigma_n^i)^{2}\delta_{\bt_1,\bt_2} \nonumber 
\end{eqnarray}
for $\bt_1$ and $\bt_2 \in \bf{T}$, where $d_T(\bt_1,\bt_2)$ denotes the patristic distance (that is, the distance in ${\bf T}$) between $\bt_1$ and $\bt_2$. This covariance function is developed in section IIC2 of \cite{JM12}: in the present context, $\sigma_f^i$ quantifies the intensity of inherited variation, $\lambda^i$ is the characteristic length-scale of the evolutionary dynamics \cite{Rasmussen06} (equivalent to the inverse of the strength of selection), and $\sigma_n^i$ quantifies the intensity of specific variation (i.e. variation unattributable to the phylogeny). We selected the hyperparameters in table \ref{Tab1}, to give different qualities to each of the three components of variation. In particular, component 2 (figure \ref{FIG1}B, upper-left, dashed line) has no inherited variation and it follows that the characteristic length-scale/strength-of-selection parameter $\lambda^2$ has no meaning for this component.

\begin{table}[!ht]
\begin{center}
\begin{tabular}{cccc}
   $i$     &  $\sigma_{f}^i$ &  $\lambda^i$  &  $\sigma_{n}^i$ \\ \hline
 1 & 4.5 & 17.9 & 0.45 \\
 2 & 0 & NA & 1 \\
 3 &  3.0 & 8.95 & 0.45 \\
\end{tabular}
\end{center}
\caption{Hyperparameter values used to generate the function-valued data shown in figure \ref{FIG1}A. The three hyperparameters (inherited variation $\sigma_{f}$, length-scale  $\lambda$ and  specific variation $\sigma_{n}$, see text), define the covariance of the basis functions (figure \ref{FIG1}B, upper-left) at different points in the tree according to equation \eqref{oucov}.}
\label{Tab1}
\end{table} 

 Each node in the tree ${\bf t} \in {\bf T}$ thus had an associated vector $w_{\bf t}=(w_{\bf t}^1,w_{\bf t}^2,w_{\bf t}^3)$ giving the weights for each of the three basis functions. These weighted basis functions were used to produce a single function-valued trait $f_{\bf t}$ at each node (of which four are shown in figure \ref{FIG1}B, lower-left panel):
\begin{equation} f_{\bf t} = w_{\bf t}^T \phi \label{basrep}
\end{equation}  where $\phi$ is the $3 \times 1024$ matrix having each $\phi_i$ as its rows.  
The resulting set of 255 curves $f_{\bf t}$ $({\bf t} \in {\bf T})$ was divided into two parts: the 128 curves at tips of ${\bf T}$ to be used as training data (that is, inputs to our regression analysis), and the 127 curves at internal nodes of ${\bf T}$ to be used for validation of the method.
 
\subsection{Dimension reduction and source separation for functional data}
\label{sec:DimensionRed} 
Since the space of all (univariate) continuous functions is infinite-dimensional, if the sample curves lie``close'' to a finite dimensional linear subspace, an acceptable approximation can be obtained by utilising weighted sums of basis curves that span that linear subspace. Effectively, this involves reversing equation \eqref{basrep}: given the curves $f_{\bf t}$, the task is to estimate the common basis $\phi$ and the weights $w_{\bf t}$. This is the challenge of `source separation' \cite{Hyvarinen:2000:ICA:351654.351659}.
 
 A more rigorous justification for this heuristic is provided by \cite{JM12}. It is shown there that a wide class of Gaussian processes of function-valued traits on a phylogeny have {\em exactly} the linear decomposition in equation \eqref{basrep}, where for each of the $i$ basis functions the weights $w^i_{\bf t}$ themselves form a (univariate) phylogenetic Gaussian process $w^i$ on the phylogeny ${\bf T}$. The task, then, is to estimate the basis $\phi$ and weights $w_{\bf t}$ at the tips of ${\bf T}$ accurately. 
 The standard approach to choosing basis curves is PCA \cite{Minka00}, which does so by explaining the greatest possible variation in successive orthogonal components, an approach extended to take account of phylogenetic relationships \cite{Revell09}. However, if a sample of functions is generated by mixing non-orthogonal basis functions, the principal components of the sample (whether or not they account for phylogeny) will not equal the basis curves, due to the assumption of orthogonality: see figure \ref{FIG1}B. If we remove the assumption of orthogonality, however, we must replace it with an alternative assumption in order to have the right number of mathematical constraints to perform source separation. In independent components analysis (ICA), the alternative assumption made is that the weight components $w^i$ are statistically independent for different values of $i$. This assumption fits naturally into our model.

PCA \textit{is} an appropriate tool for estimating true dimensionality. Therefore, to achieve both dimension reduction and source separation, we first applied PCA to the training data (the 128 curves at the tips of ${\bf T}$) to determine the appropriate value for $k$ \cite{Minka00}, which was correctly returned as $k=3$. The principal components were then passed to the `CubICA' implementation of ICA \cite{Blaschke04}. ICA has proved fruitful in other biological applications \cite{Scholz04} as has passing the results of PCA to ICA, which has been termed IPCA \cite{Yao12}. 
CubICA returned a new set of $k$ components (figure \ref{FIG1}B, lower-right panel) and, for each $i$, a corresponding weight $w^i_\bt$ at each tip $\bt$.  
An independent, univariate phylogenetic Gaussian process regression was then performed for each value of $i$, as described in \ref{sec:PhyloGP}, to obtain posterior distributions for the weights throughout the tree.  
The posterior distributions at every node in the tree (one posterior for each $w^i$) were then mapped to a single functional posterior distribution.

Using IPCA we can therefore approximately reconstruct both the basis $\phi$ and the mixing vectors $w_\bt$ from the tip data, in such as way as to be fully compatible with the phylogenetic Gaussian process framework of  \cite{JM12}.   

\subsection{Phylogenetic Gaussian process regression}
\label{sec:PhyloGP}

The mechanics of phylogenetic Gaussian process regression using a basis  $\phi$ are discussed in detail in \cite{JM12}, section IIC2. As discussed there, the phylogenetic OU process is the only stationary and Markovian Gaussian process.
For each $i=1,2,3$, under these assumptions we would therefore choose the covariance function given in equation \eqref{oucov}, and only the hyperparameters $\theta_i=(\sigma_f^i,\sigma_n^i,\lambda^i)$ would remain to be specified  
(for a fixed phylogeny ${\bf T}$). 

In the ancestral inference described in \ref{sec:DimensionRed}, we assume the parameter values in table \ref{Tab1} to be known. For the task of parameter estimation we performed the following procedure. For each of the three components $\sigma_f^i,\sigma_n^i,\lambda^i$ in turn, given the values of of the other two components 
we maximise the likelihood expression (2) in \cite{JM12} with respect to the third component to obtain their maximum likelihood estimates, and compare these to the true value.

\section{Results and Discussion}
\label{sec:ResDes}

\begin{figure*}[!ht]
\includegraphics[width=.95\textwidth]{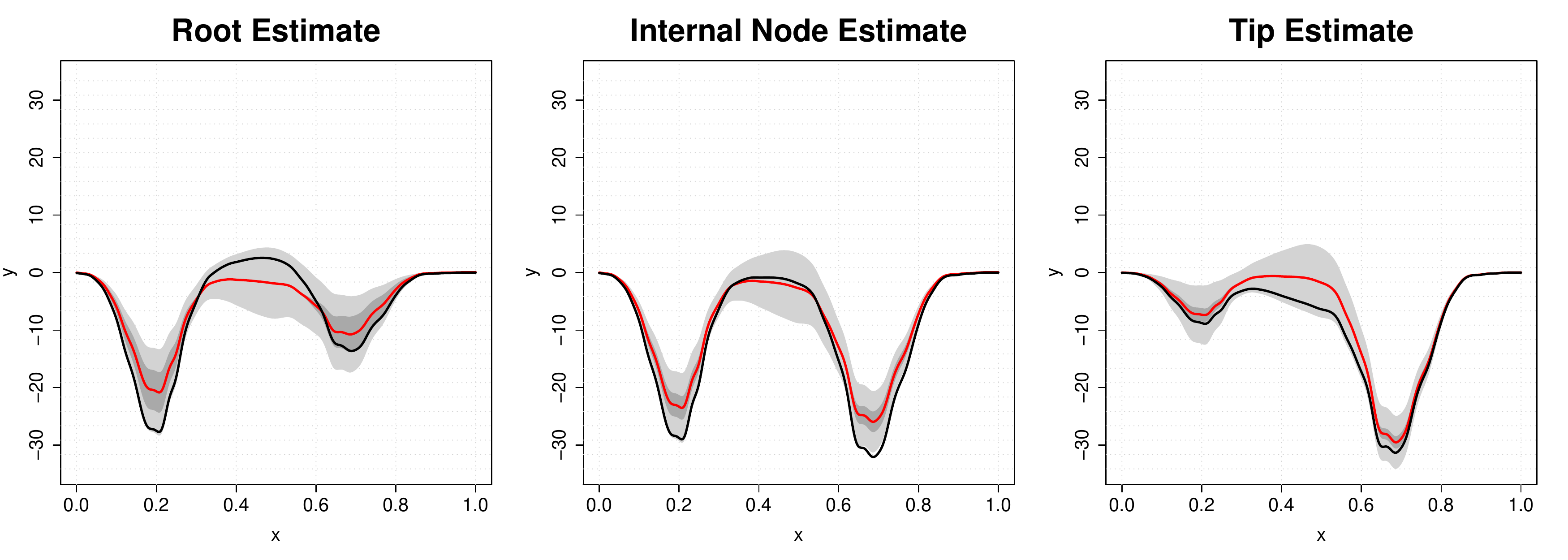}
\caption{Posterior distributions at three points in the phylogeny. The prediction made by the regression analysis shown via the posterior mean (red line), the component of posterior standard deviation due to inherited variation (dark grey band) and specific variation (light grey band). The black line shows the simulated data. In the left and centre panels this black line enables validation of the ancestral predictions. In the right panel, the black line is the training data at that tip and the posterior distribution should be understood as a prediction for an independent organism at the same point in the phylogeny. The root and internal node here are the same as those indicated in figure \ref{FIG1}A, and the tip is the second from bottom on the right of figure \ref{FIG1}A.} \label{FIG2}
\end{figure*} 
Given a tree\textbf{ T} and functional data associated with each of its tips, we shall make inferences about ancestral traits and evolutionary parameters. We simulate data on the tree in figure \ref{FIG1}A as described in \ref{sec:Simulation},
  and we estimate the independently varying basis functions $\phi_i$ using IPCA (figure \ref{FIG1}B) on tip data alone. Using the basis $\phi$ and the process hyperparameters, we derive posterior distributions over functional traits throughout the tree. Examples of the posterior distributions obtained are shown in figure \ref{FIG2}. Since specific variation (represented by the light grey band) is modelled as statistically independent at each point in $\bf{T}$, the specific variance  may simply be subtracted from the total posterior variance to obtain the posterior variance due to inherited variation (whose standard deviation is given by the width of the dark grey band). The simulated validation data is also shown in black, and can be seen typically to lie within the posterior standard deviation (given by the width of the dark grey plus light grey band). Note that the contribution of specific variation to the posterior variance is constant across the tree since tip data are silent concerning the specific component of the function-valued trait at internal nodes. On the other hand, the darker grey band decreases in width going away from the root, reflecting the decreasing (although not entirely vanishing) uncertainty concerning the inherited component of the function-valued trait as we approach the measured traits at the tips. We note that the posterior distribution, even at the root, puts a clear constraint on the possible ancestral functional data: in this (admittedly simulated and highly controlled) setting we can reason effectively about ancestral function-valued traits. 

We next considered whether parameters describing evolutionary dynamics could be estimated, given only functional data from the tips of the tree. Specifically, we fixed two of the three components of $\theta_i$ to random values and estimated the third: results shown in figures S1-2. 
The accuracy of the predictions is very encouraging: essentially they are problematic only on relatively rare occasions when an over-simplified model is fitted excluding either inherited or specific variation entirely. Similarly promising statistical performance was obtained when  $\lambda$ was known 
and knowledge of either $\sigma_f$ or $\sigma_n$ was replaced by knowledge of the ratio between them.
 
When there is greater uncertainty over the hyperparameters, we anticipate that the use of hyperpriors will enable statistical performance to be maintained for inference about ancestral function-valued traits, but we reserve this for future work.

R Code for the IPCA, ancestral reconstruction and hyperparameter estimation is available from http://tinyurl.com/FuncPhylo1

\bibliography{Bibliography}
\bibliographystyle{vancouver}
 
\end{document}